\documentclass{article}

     \usepackage[preprint,nonatbib]{neurips_2023}

\usepackage[utf8]{inputenc} 
\usepackage[T1]{fontenc}    
\usepackage{hyperref}       
\usepackage{url}            
\usepackage{booktabs}       
\usepackage{amsfonts}       
\usepackage{nicefrac}       
\usepackage{microtype}      
\usepackage{xcolor}         

\usepackage{multirow}
\usepackage{booktabs}
\usepackage{colortbl} 
\usepackage{enumitem}
\usepackage{graphicx}
\usepackage{wrapfig,lipsum,booktabs}
\newcommand{\etal}{\textit{et al.}}

\title{Symmetry Defense Against \\ XGBoost Adversarial Perturbation Attacks}

\author{%
  Blerta Lindqvist
  Department of Computer Science\\
  Aalto University\\
  Espoo, Finland \\
  \texttt{blerta.lindqvist@aalto.fi} \\
}

\author{Blerta Lindqvist\\
Aalto University\\
{\tt\small }
 \texttt{blerta.lindqvist@aalto.fi} \\
}

\begin{document}

\maketitle

\begin{abstract}
We examine whether symmetry can be used to defend tree-based ensemble classifiers such as gradient-boosting decision trees (GBDTs) against adversarial perturbation attacks. The idea is based on a recent symmetry defense for convolutional neural network classifiers (CNNs) that utilizes CNNs' lack of invariance with respect to symmetries. CNNs lack invariance because they can classify a symmetric sample, such as a horizontally flipped image, differently from the original sample. CNNs' lack of invariance also means that CNNs can classify symmetric adversarial samples differently from the incorrect classification of adversarial samples. Using CNNs' lack of invariance, the recent CNN symmetry defense has shown that the classification of symmetric adversarial samples reverts to the correct sample classification.
In order to apply the same symmetry defense to GBDTs, we examine GBDT invariance and are the first to show that GBDTs also lack invariance with respect to symmetries. We apply and evaluate the GBDT symmetry defense for nine datasets against six perturbation attacks with a threat model that ranges from zero-knowledge to perfect-knowledge adversaries.
Using the feature inversion symmetry against zero-knowledge adversaries, we achieve up to 100\% accuracy on adversarial samples even when default and robust classifiers have 0\% accuracy.
Using the feature inversion and horizontal flip symmetries against perfect-knowledge adversaries, we achieve up to over 95\% accuracy on adversarial samples for the GBDT classifier of the F-MNIST dataset even when default and robust classifiers have 0\% accuracy.
\end{abstract}

\section{Introduction}

Tree-based ensemble classifiers such as gradient-boosted decision trees (GBDTs) are popular classifiers~\cite{andriushchenko2019provably} that are susceptible to adversarial perturbation attacks~\cite{kantchelian2016evasion,cheng2018queryefficient,andriushchenko2019provably,cheng2019sign,zhang2020efficient,chen2020hopskipjumpattack}. 
The popularity of GBDTs is due to their interpretability, performance, and efficient implementations~\cite{chen2016xgboost,ke2017lightgbm}.
Adversarial perturbation attacks, which perturb original samples imperceptibly to cause misclassification, were first discovered in convolutional neural networks (CNNs)~\cite{szegedy2013intriguing,goodfellow6572explaining,moosavi2016deepfool,carlini2017towards,madry2017towards}.
Some of the strongest CNN perturbation attacks are gradient-based attacks~\cite{moosavi2016deepfool,madry2017towards,carlini2017towards}, which do not apply to GBDTs because GBDTs are non-continuous step functions that lack a gradient.
Black-box attacks have been applied successfully on both CNNs and GBDTs~\cite{chen2017zoo,cheng2018queryefficient,cheng2019sign,chen2020hopskipjumpattack}.
Black-box, gradient-based attacks~\cite{chen2017zoo,ilyas2018black} approximate the gradient of an XGBoost classifier given that XGBoost classifiers are non-continuous step functions, therefore, non-differentiable.
Decision-based black-box attacks~\cite{brendel2017decision,cheng2018queryefficient,cheng2019sign,chen2020hopskipjumpattack}, otherwise known as hard-label black-box attacks, can also be applied to XGBoost classifiers. Decision-based attacks start from an adversarial sample, then minimize the perturbation while remaining close to the classifier boundary.
Other attacks, such as MILP~\cite{kantchelian2016evasion} and LT-Attack~\cite{zhang2020efficient}, are customized for GBDTs and utilize the tree structure of GBDTs classifiers~\cite{kantchelian2016evasion,zhang2020efficient}.

Current GBDT defenses against adversarial attacks focus on robustness. Similarly to CNN adversarial training (AT)~\cite{madry2017towards}, adversarial boosting~\cite{kantchelian2016evasion} needs attack knowledge to generate adversarial samples for training. Other GBDT defenses against adversarial perturbation attacks use different approaches to achieve robustness:
training with adversarial samples~\cite{kantchelian2016evasion}, using domain knowledge to increase attack cost~\cite{chen2021cost}, robustness increase ~\cite{andriushchenko2019provably,chen2019robust,Chen2019TrainingRT}, or using a loss function~\cite{Calzavara2019TreantTE}.


\begin{figure}
\begin{minipage}[c]{0.49\columnwidth}
  \centering
  \includegraphics[width=1.00\columnwidth]{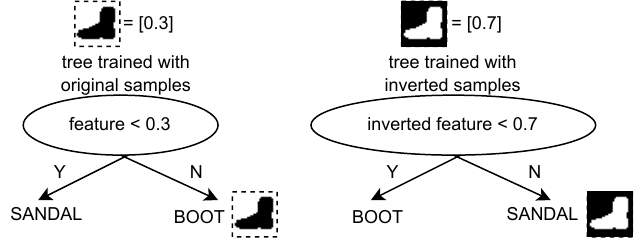}
  \caption{Lack of invariance in GBDTs. Although the splitting conditions correspond exactly, the boot sample classifies differently in these toy examples of 1-tree GBDT classifiers trained separately on 1-feature original and inverted samples.}
  \label{fig:imbalance}
\end{minipage}
\hfill
\begin{minipage}[c]{0.49\columnwidth}
  \centering
  \includegraphics[width=1.00\columnwidth]{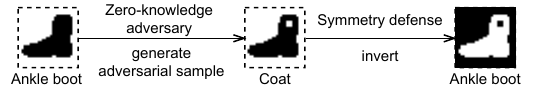}
  \caption{Here, we illustrate the symmetry defense against zero-knowledge adversaries. The classifier is trained with both original and inverted samples. The symmetry defense inverts and then classifies the adversarial sample. The defense processes original samples the same way.}
  \label{fig:xgboost}
\end{minipage}
\end{figure}



A recent symmetry defense~\cite{lindqvist2022symmetry} against CNN perturbation attacks achieves close-to-default accuracies against adversarial attacks without using any attack knowledge. The CNN symmetry defense utilizes the CNN lack of invariance with respect to symmetries~\cite{azulay2019deep,bouchacourt2021grounding,engstrom2019exploring,kayhan2020translation}, which means that CNNs can classify symmetric samples differently. 
The CNN lack of invariance also means that CNNs can classify symmetric adversarial samples differently from the classification of adversarial samples.
By applying symmetry to adversarial samples, the symmetry defense reverts the classification of adversarial samples to the original classification, even against adversaries aware of the defense.
Based on the CNN symmetry defense~\cite{lindqvist2022symmetry}, we pose the question:

\emph{Can the CNN symmetry defense be applied to tree-based ensemble classifiers such as GBDTs?}

In this paper, we address this question with the following main contributions:

\begin{itemize} [leftmargin=*]
  \setlength\itemsep{0em}
  
  \item We are the first to show that XGBoost classifiers of all considered datasets lack invariance to symmetries as CNNs do.
  
  \item We identify the imbalance of XGBoost splitting conditions as an inherent cause of XGBoost lack of invariance even if the splitting condition values correspond. We show this in two 1-tree classifiers trained respectively with original and inverted samples shown in Figure~\ref{fig:imbalance}. Furthermore, the XGBoost algorithm for greedy split finding and other XGBoost design choices can also contribute to the lack of invariance by causing non-symmetric split values in trees in symmetric settings, as discussed in Section~\ref{sec:discussion}.
  
  \item Using symmetry defense and no attack knowledge, we defend against six state-of-the-art attacks from adversaries that range from zero-knowledge to perfect-knowledge.
  
  \item Using the feature inversion symmetry even in datasets that lack inherent symmetries, we defend GBDT classifiers against zero-knowledge adversaries, exceeding default and robust symmetries by up to 100\% points on adversarial samples. The symmetry defense against zero-knowledge adversaries is illustrated in Figure~\ref{fig:xgboost} and experimental results are shown in Table~\ref{table:zero_attacks_test_linf}, and Table~\ref{table:zero_attacks_test_l2} in Appendix~\ref{sec:zero_attacks_test_l2}.
  
  \item Using the invert and flip symmetries, we defend the GBDT classifier of F-MNIST against perfect-knowledge adversaries, exceeding default and robust classifiers by up to over 95\%, as shown in Table~\ref{table:perfect_attacks_linf} and in Table~\ref{table:perfect_attacks_l2} in Appendix~\ref{sec:prefect_l2}.The defense is illustrated in Figure~\ref{fig:perfect} in Appendix~\ref{sec:perfect}.
  
  \item We observe that the MILP, LT-Attack and Cube attacks are largely unable to generate adversarial samples against XGBoost models against zero-knowledge adversaries that are trained with original and inverted samples. The adversarial samples generated by these attacks are non-adversarial even if the samples are not inverted before classification as shown in Figure~\ref{fig:xgboost}. We show the experimental results in Table~\ref{table:zero_attacks_no_inversion_linf} and discuss them in Section~\ref{sec:discussion}.
  
  \item We find that augmentation of the training dataset with legitimate symmetric samples affects classifier adversarial robustness, as discussed in Section~\ref{sec:discussion}.
  
  
\end{itemize}

\section{Background and Related Work}

\subsection{GBDT Classifiers}

The widely-used GBDT classifiers~\cite{andriushchenko2019provably} are non-continuous step-functions that lack a gradient. GBDTs use $K$ additive functions to predict the output. A GBDT classifier consists of $K$ decision trees, where each $t$ tree is a weak learner. Each non-leaf node branches out to two children in a tree based on a splitting condition. GBDTs build trees greedily by making locally optimal splitting conditions. The path from the tree root is determined by the splitting conditions that the sample satisfies or not. The final leaf node of the path has a score value that determines the score value of the sample for that tree. The score values of individual trees are used to calculate the final prediction as the sum of the scores of the corresponding leaves in all the trees. A GBDT ensemble classifier is optimized by iteratively adding a tree that minimizes the error of the previous trees.

XGBoost~\cite{chen2016xgboost} is a widely-used GBDT classifier with state-of-the-art results~\cite{ke2017lightgbm,zhang2020efficient}. Unable to enumerate all possible tree structures, XGBoost~\cite{chen2016xgboost} uses a greedy algorithm that starts from a single leaf and adds branches iteratively. The basic exact split finding algorithm of XGBoost chooses the best split among all the possible splits on all features, which is computationally demanding, especially for continuous features. For non-exact split finding, XGBoost finds candidate splitting points for features based on feature distribution percentiles. Split candidates can be calculated at the beginning of the tree construction but can also be refined after splits.


\subsection{Symmetry, Equivariance and Invariance}

Symmetry is a transformation that leaves an object invariant~\cite{dummit2004abstract,schmidt2012learning}, for example, the horizontal flipping of an image. Equivariance and invariance are used to characterize functions that take as input objects on which symmetry can be applied. A function is \emph{equivariant} with respect to a symmetry transformation if that transformation can be equally applied to the function's input or output. emph{Invariance} is a particular case of equivariance where we apply the identity transformation on the function output. For classifier functions that output a label for a sample input, invariance means that symmetric samples have the same output. Mathematical definitions of equivariance, invariance, and a discussion of symmetry groups are in Appendix~\ref{sec:symmetry}.

\subsubsection{CNN Symmetry Defense}

Lindqvist~\cite{lindqvist2022symmetry} has recently proposed a CNN Symmetry Defense (SD) against adversarial attacks, based on CNNs' lack of invariance with respect to symmetries~\cite{azulay2019deep,bouchacourt2021grounding,engstrom2019exploring,kayhan2020translation}. Lindqvist~\cite{lindqvist2022symmetry} utilizes CNN lack of invariance to revert the classification of adversarial samples to the original correct classification for several attacks without using any attack knowledge.
The SD threat model is based on~\cite{carlini2019evaluating} and ranges from zero-knowledge adversaries to perfect-knowledge adversaries.


Against zero-knowledge adversaries, SD uses the invert or flip symmetry. One version of SD trains a classifier with both original and flipped samples. It classifies a sample by applying symmetry to it and then classifying it, as illustrated in Figure~\ref{fig:xgboost}. Another version of SD trains two separate classifiers, one with original samples and one with symmetric samples. Then SD classifies samples with the classifier trained with symmetric samples, of which zero-knowledge adversaries are unaware.

Against perfect-knowledge adversaries, SD uses the invert and flip symmetries to construct a symmetry group. To classify a sample, SD first applies all four symmetry group symmetry transformations to the sample because adversaries aware of the defense can apply any subgroup symmetries before adversarial generation. Then, SD classifies all these samples and assigns a label when at least two of their classification labels agree. Notably, the closure of the symmetry group ensures that any combination of consecutive symmetry transformations is the same as one of the symmetry group transformations. Figure~\ref{fig:perfect} in Appendix~\ref{sec:perfect} illustrates the symmetry group defense.


\subsection{Adversarial Attacks on GBDTs} 


Gradient-based attack methods used in CNNs do not apply to GBDTs because GBDTs are non-continuous step functions. Instead, we consider several adversarial attacks that are specifically targeted at GBDT classifiers~\cite{kantchelian2016evasion,andriushchenko2019provably,zhang2020efficient}, as well as several black-box attacks~\cite{cheng2018queryefficient,cheng2019sign,chen2020hopskipjumpattack}. 
The mixed-integer linear programming (\textbf{MILP}) attack~\cite{kantchelian2016evasion} is a white-box attack that finds the exact value of minimal perturbation. However, the disadvantage of MILP is that it has NP-complete complexity and can be slow for large-scale models.
The \textbf{LT-Attack}~\cite{zhang2020efficient} formulates the attack problem as a discrete search problem for tree ensembles and solves it with a simple, greedy algorithm. The LT-Attack~\cite{zhang2020efficient} scales well with even hundreds of trees and is thousands of times faster than the MILP attack. The LT-Attack aims to find the smallest adversarial perturbation and finds smaller perturbations than other black-box, decision-based attacks.
The \textbf{Cube} attack~\cite{andriushchenko2019provably} is a simple attack inspired by random search, is based on an evolutionary algorithm, and lacks theoretical guarantees. The Cube attack changes a random subset of coordinates at every iteration and accepts the change if the functional margin decreases. The Cube attack only supports $L_\infty$ perturbations. Therefore, we report the $L_2$ perturbation of the $L_\infty$ attacks, similarly to~\cite{zhang2020efficient}.
The \textbf{OPT} attack~\cite{cheng2018queryefficient} is a hard-label black-box attack that reformulates finding adversarial samples as a continuous optimization problem solvable by zeroth order optimization.
The \textbf{SignOPT} attack~\cite{cheng2019sign} is a decision-based attack that uses hard-label, black-box queries.
The Hopskipjumpattack (\textbf{HSJA}) attack~\cite{chen2020hopskipjumpattack} is a decision-based black-box attack that is based only on the output labels.

\subsection{Adversarial Defenses for GBDTs} 

\textbf{Adversarial Robustness.} Robustness is defined as the minimum value of the adversarial perturbation~\cite{zhang2020efficient}. According to this definition, the higher the minimum adversarial perturbation is, the more robust the classifier is.
Finding the exact value of the minimal adversarial perturbation requires exponential time and is NP-complete~\cite{kantchelian2016evasion}.
\cite{andriushchenko2019provably} showed that adversarial robustness of GBDTs of depth $1$ has polynomial time.
\cite{chen2019robustness} showed a polynomial time algorithm for verifying the robustness of a single tree.
Without any formal guarantees, an approximation of the inner minimization of the max-min saddle point problem under worst-case perturbation is incorporated into the tree-building algorithm by~\cite{chen2019robust}.
\cite{kantchelian2016evasion} find the minimal adversarial perturbation with a mixed integer linear program that does not scale well. Kantchelian~\etal~\cite{kantchelian2016evasion} also find that training with additional samples decreases the adversarial robustness.
Zhang~\etal~\cite{zhang2020efficient} discretize the input space and implement a simple greedy algorithm up to thousands of times faster than~\cite{kantchelian2016evasion}.

\textbf{Adversarial Boosting.} Kantchelian~\etal~\cite{kantchelian2016evasion} use adversarial boosting to improve the robustness of GBDT classifiers. Adversarial boosting augments the training dataset with adversarial samples during the training. During a boosting round, they generate adversarial samples using the current model and the original samples of the training dataset. The classifier is then trained on both original and adversarial samples. Adversarial boosting is similar to the CNN AT defense~\cite{kurakin2016adversarial,madry2017towards,szegedy2013intriguing} against adversarial perturbation attacks.
Importantly, Kantchelian~\etal~\cite{kantchelian2016evasion} find that training with additional samples decreases adversarial robustness.

\textbf{Cost-Aware Robust Decision Trees.} Chen~\etal~\cite{chen2021cost} utilize domain knowledge of the asymmetric costs of feature manipulation to increase the cost for adversarial perturbation attack. Furthermore, they use feature manipulation cost as a constraint in node splitting.

\textbf{Robust Decision Trees.} Chen~\etal~\cite{chen2019robust} aim to optimize classifier performance for worst-case adversarial perturbation with a max-min saddle point problem that is used in the tree building. To make this problem tractable, they approximate the inner minimization. As a result, Chen~\etal~\cite{chen2019robust} increase the minimal adversarial perturbation, that is, adversarial robustness.

\textbf{Provably Robust Boosting.} Provably robust boosting~\cite{andriushchenko2019provably} calculate efficiently and then minimize the upper bound of the robust test error as the sum of the maximum losses of each tree.

\textbf{Training Robust Tree Ensembles.} To increase robustness, Chen~\etal~\cite{Chen2019TrainingRT} use a greedy heuristic to approximate the saddle point objective.

\textbf{TREANT.} Calzavara~\etal~\cite{Calzavara2019TreantTE} build trees greedily not by using a splitting condition, but by optimizing an evasion-aware loss function.



\subsection{Adversarial Attacks on CNNs} 

Adversarial perturbation attacks apply imperceptible perturbations to samples in order to cause misclassification~\cite{carlini2017towards,goodfellow6572explaining,madry2017towards,szegedy2013intriguing}. Attacks can be divided into white-box attacks and black-box attacks. In white-box attacks, the adversary knows and can use the classifier model and parameters and can use them for the attack. White-box attacks include the FGSM~\cite{goodfellow6572explaining}, Carlini-Wagner~\cite{carlini2017towards}, PGD~\cite{madry2017towards}, Auto-PGD~\cite{croce2020reliable} attacks. White-box attacks generally use the classifier gradient to construct adversarial samples. In black-box attacks, the classifier is like a black box where the classifier model and parameters are unknown, but the adversary can make queries to obtain classifier score or probability outputs. Black-box attacks include the ZOO attack~\cite{chen2017zoo}, the query-efficient OPT attack by~\cite{cheng2018queryefficient}, the Sign-OPT attack~\cite{cheng2019sign}, the Square attack~\cite{andriushchenko2020square}. The ZOO attack~\cite{chen2017zoo} uses classifier probability outputs, whereas the query-efficient attack~\cite{cheng2018queryefficient} uses classifier score outputs.

\subsection{Adversarial Defenses for CNNs} 

\textbf{Adversarial Training.} Adversarial training (AT)~\cite{kurakin2016adversarial,madry2017towards,szegedy2013intriguing} uses adversarial perturbation samples to train the classifier, making the classifier more robust to adversarial samples. Madry~\etal~\cite{madry2017towards} have introduced the robust PGD AT, where they formulate the defense as a min-max optimization problem. However, AT reliance on adversarial samples makes AT inapplicable when the attack is unknown. In addition, generating adversarial samples for training with AT increases training time and computation.

\textbf{Failed Defenses.} Many other defenses have been shown to fail against an adaptive adversary. For example, defensive distillation has been shown to be not robust to adversarial samples~\cite{carlini2016defensive}, many adversarial detection defenses have been bypassed~\cite{carlini2017adversarial,carlini2017magnet}, obfuscated gradient defenses~\cite{athalye2018obfuscated} and other defenses~\cite{tramer2020adaptive} have been circumvented.

\textbf{Other Rejected Defenses.} Many CNN defenses have failed to defend against perfect-knowledge adversaries that can adapt. These include defensive distillation defenses~\cite{carlini2016defensive}, many bypassed adversarial detection defenses~\cite{carlini2017magnet,carlini2017adversarial}, several obfuscated gradient defenses~\cite{athalye2018obfuscated}.

The recent SD defense~\cite{lindqvist2022symmetry} uses CNNs' lack of invariance to counter adversarial attacks.

\textbf{Summary.} Relevant to this paper, GBDT invariance with respect to symmetries has not been examined previously, and there is no previous GBDT symmetry defense.

\section{Experimental Settings}

The experimental setup is based on~\cite{zhang2020efficient}. In addition to the datasets used in~\cite{zhang2020efficient}, we use datasets with samples that are symmetric to samples of original datasets. We conduct experiments for both $L_2$ and $L_\infty$ attack perturbations. The MILP implementation uses the Gurobi Solver~\cite{gurobi2021gurobi}.


\textbf{Evaluation Metrics.} We use both accuracies and the mean adversarial perturbation values for evaluation.
\emph{Default accuracy} is calculated from the number of samples that classify correctly out of all samples. \emph{Adversarial accuracy} is calculated from the number of adversarial samples that classify correctly out of all samples that classify correctly (the adversarial samples are generated from original samples that classify correctly). \emph{Adversarial perturbation} is calculated as the mean adversarial perturbation value of adversarial samples that misclassify but are classified correctly as original samples.


\textbf{Datasets.} We evaluate the symmetry defense on nine public, binary, and multi-classification datasets also used by~\cite{chen2019robustness,zhang2020efficient}: breast-cancer~\cite{Dua_2019}, diabetes~\cite{smith1988using}, MNIST2-6~\cite{lecun1998mnist}, ijcnn~\cite{prokhorov2001ijcnn}, MNIST~\cite{lecun1998mnist}, F-MNIST~\cite{xiao2017fashion}, webspam~\cite{wang2012evolutionary}, covtype~\cite{Dua_2019}, HIGGS~\cite{baldi2014searching}.
We train the models with standard GBDT models with the same training parameters as in~\cite{chen2019robustness,zhang2020efficient}, shown in Appendix~\ref{app:training_params}. Datasets features are normalized to the $[0,1]$ range. We list datasets in tables in order of increasing training data size.

\textbf{Inverted and Flipped Datasets.} From original datasets, we derive datasets with samples that are symmetric to samples from original datasets. We use the inverted symmetry for all datasets due to the lack of inherent symmetries for most considered datasets, converting each $a$ feature value to a $1-a$ value. We assemble datasets that contain original and inverted samples for each original dataset. For the F-MNIST dataset, we also assemble a dataset that contains original, inverted, flipped, and flipped and inverted F-MNIST samples.

\textbf{Model Training and Parameters.} Default, robust, and symmetry defense XGBoost models were all trained with the parameters shown in Table~\ref{table:params} in Appendix~\ref{app:training_params}, defined in~\cite{chen2019robustness} and also used in~\cite{zhang2020efficient}. The trained robust models are obtained from the robust GBDT training~\cite{chen2019robustness}. The symmetry defense models against zero-knowledge adversaries are trained with original and inverted samples. The symmetry defense model against zero-knowledge adversaries for F-MNIST trains the classifier with original, inverted, flipped, and inverted and flipped samples.

\textbf{Attacks.} We evaluate against six attacks with same attack settings as in~\cite{zhang2020efficient}: MILP~\cite{kantchelian2016evasion}, LT-Attack~\cite{zhang2020efficient}, Cube~\cite{andriushchenko2019provably}, OPT~\cite{cheng2018queryefficient}, SignOPT~\cite{cheng2019sign}, HSJA~\cite{chen2020hopskipjumpattack}. We run all experiments with $20$ threads per task. We evaluate attacks on the first $500$ testing samples (or the entire testing dataset when smaller than $500$), or on the first $50$ (marked with an *) when attacks run long.

\textbf{Computational resources.} The defense doubles or quadruples computational resources because of augmenting the training dataset. We discuss computational resources in more detail in Appendix~\ref{sec:computation}.


\subsection{Threat Model} 

Our threat model is the same as the one defined in~\cite{lindqvist2022symmetry}, based on~\cite{carlini2019evaluating}. We assume that the attacker knows the model and its parameters and consider the following adversaries:

\begin{itemize} 
  \setlength\itemsep{0em}
  \item \textbf{Zero-Knowledge.} The adversary is not aware of the defense.
  \item \textbf{Perfect-Knowledge.} The adversary is aware of the defense and adapts its attack based on the defense.
  \item \textbf{Limited-Knowledge.} According to~\cite{carlini2019evaluating} recommendations, limited-knowledge adversaries need to be evaluated only when zero-knowledge adversaries fail and perfect-knowledge adversaries succeed. The proposed symmetry defense succeeds in both cases.
\end{itemize}

\section{XGBoost Classifiers Lack Invariance with Respect to Symmetries} \label{sec:gbdt_equi} 

Here, we examine XGBoost lack of invariance with respect to the invert symmetry in order to determine whether symmetry can be used for a GBDT symmetry defense that is similar to the recent CNN symmetry defense by Lindqvist~\cite{lindqvist2022symmetry} against adversarial perturbation attacks. For each dataset, we train a classifier with both original and inverted samples with default parameter settings based on Appendix~\ref{app:training_params}. We test with original and inverted testing samples, aiming to find pairs of corresponding original and inverted samples that are classified differently.

We are the first to show that XGBoost classifiers lack invariance with respect to the inversion symmetry. Table~\ref{table:equivariance} in Appendix~\ref{sec:invariance} shows that an XGBoost classifier trained with both original and inverted samples does not always classify an original sample and its inverted symmetry sample the same.

\section{Proposed Symmetry Defense For XGBoost Classifiers}


The symmetry defense for GBDT classifiers uses the same symmetries as the symmetry defense for CNNs~\cite{lindqvist2022symmetry}. Against zero-knowledge adversaries, we use the feature invert symmetry since most considered datasets lack inherent symmetries. Against perfect-knowledge adversaries of the F-MNIST classifier, we use the invert and the flip symmetry for the symmetry group.

\subsection{Symmetry Defense Against Zero-Knowledge Adversaries}

The symmetry defense against zero-knowledge adversaries uses the invert symmetry, training XGBoost classifiers with both original and inverted samples. The defense processes original and adversarial samples the same, inverting samples before classification, as shown in Figure~\ref{fig:xgboost}. The defense exceeds default and robust classifier accuracies by up to 100\% points. The experimental results are in Table~\ref{table:zero_attacks_test_linf} for $L_\infty$ perturbation attacks, and in Table~\ref{table:zero_attacks_test_l2} in Appendix~\ref{sec:zero_attacks_test_l2} for $L_\infty$ perturbation attacks.

\begin{table*}[ht!]
\caption{The proposed symmetry defense exceeds the accuracies of default and robust classifiers against $L_\infty$ perturbation attacks from zero-knowledge adversaries. 
Adversarial accuracies and perturbation are calculated from samples that classify correctly as original samples.
Adversarial accuracies are calculated from the number of adversarial samples that classify correctly out of all original samples that classify correctly.
Adversarial perturbation mean values are calculated from adversarial samples that misclassify that were generated from original samples that were classified correctly.
Results for $L_2$ attacks are in Table~\ref{table:zero_attacks_test_l2} in Appendix~\ref{sec:zero_attacks_test_l2}.}
\centering
\setlength{\tabcolsep}{0pt}
\begin{tabular}{@{\extracolsep{2pt}}llllrrrrrrrrrrrrr@{}} 
\toprule
\multirow{2}{*}{\rotatebox[origin=c]{90}{Data-}} & \multirow{2}{*}{\rotatebox[origin=c]{90}{set}} & \multirow{2}{*}{\rotatebox[origin=c]{90}{Cl.}} & Def. && \multicolumn{12}{c}{Adversarial accuracy and perturbation} \\
\cmidrule{5-16}
&&& acc. & \multicolumn{2}{c}{MILP} & \multicolumn{2}{c}{LT-Attack} & \multicolumn{2}{c}{Cube} & \multicolumn{2}{c}{OPT} & \multicolumn{2}{c}{SIGNOPT} & \multicolumn{2}{c}{HSJA} \\
\midrule
\multirow{3}{*}{\rotatebox[origin=c]{90}{breast-}} & \multirow{3}{*}{\rotatebox[origin=c]{90}{cancer}}
&   Def.  & 86.4\% &     0.0\% & .212 &    0.0\% & .190 &    0.0\% & .338 &    0.0\% & .229 &    0.0\% & .223 &   0.0\% & .212 & \\
&  & Rob. & 87.0\% &     3.0\% & .407 &    0.0\% & .404 &    0.0\% & .750 &    0.0\% & .322 &    0.0\% & .322 &   0.0\% & .322 & \\
&  & Sym. & 87.0\% &    88.1\% & .167 &   86.0\% & .154 &   38.8\% & .341 &   62.7\% & .120 &   91.8\% & .222 &  29.9\% & .133 & \\
\cmidrule{1-3} \cmidrule{4-4} \cmidrule{5-6} \cmidrule{7-8} \cmidrule{9-10} \cmidrule{11-12} \cmidrule{13-14} \cmidrule{15-16}
\multirow{3}{*}{\rotatebox[origin=c]{90}{diabetes}} & 
&   Def.  & 87.3\% &     0.0\% & .045 &    0.0\% & .282 &    0.0\% & .074 &    0.0\% & .056 &    0.0\% & .052 &    0.0\% & .050 & \\
& & Rob.  & 90.3\% &     0.0\% & .112 &    0.0\% & .113 &    0.0\% & .373 &    0.0\% & .103 &    0.0\% & .102 &    0.0\% & .102 & \\
& & Sym.  & 86.1\% &    89.3\% & .044 &   75.0\% & .034 &   60.3\% & .091 &   59.5\% & .065 &   78.5\% & .055 &   62.8\% & .048 & \\
\cmidrule{1-3} \cmidrule{4-4} \cmidrule{5-6} \cmidrule{7-8} \cmidrule{9-10} \cmidrule{11-12} \cmidrule{13-14} \cmidrule{15-16}
\multirow{3}{*}{\rotatebox[origin=c]{90}{MNIST}} & \multirow{3}{*}{\rotatebox[origin=c]{90}{2-6}}
&   Def.  & 99.0\% &    69.3\% & .033 &    0.0\% & .083 &    0.0\% & .111 &    0.0\% & .386 &    0.0\% & .225 &    0.0\% &  .131 & \\
& & Rob.  & 99.8\% &   *94.0\% &*.310 &    0.0\% & .325 &    0.0\% & .325 &    0.0\% & .599 &    0.0\% & .421 &    0.0\% &  .352 & \\
& & Sym.  & 99.2\% &    82.9\% & .039 &   99.8\% & .002 &   73.8\% & .104 &   83.7\% & .159 &   92.9\% & .038 &   91.5\% & .007 & \\
\cmidrule{1-3} \cmidrule{4-4} \cmidrule{5-6} \cmidrule{7-8} \cmidrule{9-10} \cmidrule{11-12} \cmidrule{13-14} \cmidrule{15-16}
\multirow{3}{*}{\rotatebox[origin=c]{90}{ijcnn}} &
&   Def.  & 99.2\% &     0.0\% & .017 &    0.0\% & .018 &    0.0\% & .018 &    0.0\% & .018 &    0.0\% & .018 &    0.0\% & .018 & \\
& & Rob.  & 95.4\% &     0.0\% & .022 &    0.0\% & .023 &    0.0\% & .034 &    0.0\% & .035 &    0.0\% & .034 &    0.0\% & .034 & \\
& & Sym.  & 99.4\% &    90.5\% & .015 &   81.6\% & .016 &   46.6\% & .018 &   54.6\% & .020 &   81.7\% & .018 &   61.7\% & .018 & \\
\cmidrule{1-3} \cmidrule{4-4} \cmidrule{5-6} \cmidrule{7-8} \cmidrule{9-10} \cmidrule{11-12} \cmidrule{13-14} \cmidrule{15-16}
\multirow{3}{*}{\rotatebox[origin=c]{90}{MNIST}} &
&   Def.  & 98.2\% &    64.0\% & .003 &    0.0\% & .020 &    0.0\% & .045 &    0.0\% & .169 &    0.0\% & .085 &   *0.0\% &*.022 & \\
& & Rob.  & 98.6\% &   *80.0\% &*.304 &    0.0\% & .292 &    0.0\% & .281 &    0.0\% & .516 &    0.0\% & .362 &    0.0\% & .313 & \\
& & Sym.  & 98.0\% &    98.0\% & .005 &   99.6\% & .001 &   66.7\% & .030 &  *92.0\% & .059 &  *98.0\% & .063 & *100.0\% &*-    & \\
\cmidrule{1-3} \cmidrule{4-4} \cmidrule{5-6} \cmidrule{7-8} \cmidrule{9-10} \cmidrule{11-12} \cmidrule{13-14} \cmidrule{15-16}
\multirow{3}{*}{\rotatebox[origin=c]{90}{F-}} & \multirow{3}{*}{\rotatebox[origin=c]{90}{MNIST}}
&   Def.  & 90.4\% &   *81.0\% &*.005 &    0.0\% & .022 &    0.0\% & .036 &    0.0\% & .144 &    0.0\% & .064 &    0.0\% & .039 & \\
& & Rob.  & 91.2\% &   *97.7\% &*.104 &    0.0\% & .096 &    0.0\% & .094 &    0.0\% & .251 &    0.0\% & .138 &    0.0\% & .110 & \\
& & Sym.  & 90.4\% &    98.7\% & .003 &   96.0\% & .002 &   95.8\% & .021 &   93.8\% & .120 &   95.8\% & .032 &   93.5\% & .023 & \\
\cmidrule{1-3} \cmidrule{4-4} \cmidrule{5-6} \cmidrule{7-8} \cmidrule{9-10} \cmidrule{11-12} \cmidrule{13-14} \cmidrule{15-16}
\multirow{3}{*}{\rotatebox[origin=c]{90}{web-}} & \multirow{3}{*}{\rotatebox[origin=c]{90}{spam}}
&   Def.  & 99.0\% &    75.4\% & .000 &    0.0\% & .001 &    0.0\% & .003 &    0.0\% & .010 &    0.0\% & .004 &    0.0\% & .005 & \\
& & Rob.  & 98.4\% &    *0.0\% &*.013 &    0.0\% & .017 &    0.0\% & .036 &    0.0\% & .102 &    0.0\% & .054 &    0.0\% & .064 & \\
& & Sym.  & 99.2\% &   100.0\% & -    &   70.3\% & .000 &   84.5\% & .002 &   91.6\% & .009 &   98.0\% & .000 &   93.5\% & .004 & \\
\cmidrule{1-3} \cmidrule{4-4} \cmidrule{5-6} \cmidrule{7-8} \cmidrule{9-10} \cmidrule{11-12} \cmidrule{13-14} \cmidrule{15-16}
\multirow{3}{*}{\rotatebox[origin=c]{90}{covtype}} &
&   Def.  & 93.0\% &   *93.6\% & .013 &    0.0\% & .022 &    0.0\% & .029 &    0.0\% & .034 &    0.0\% & .029 &    0.0\% & .030 & \\
& & Rob.  & 86.0\% &   *86.7\% & .033 &    0.0\% & .048 &    0.0\% & .102 &    0.0\% & .095 &    0.0\% & .091 &    0.0\% & .091 & \\
& & Sym.  & 91.2\% &   *89.4\% &*.101 &   46.9\% & .013 &   67.9\% & .029 &   75.7\% & .041 &   81.8\% & .034 &   77.0\% & .034 & \\
\cmidrule{1-3} \cmidrule{4-4} \cmidrule{5-6} \cmidrule{7-8} \cmidrule{9-10} \cmidrule{11-12} \cmidrule{13-14} \cmidrule{15-16}
\multirow{3}{*}{\rotatebox[origin=c]{90}{HIGGS}} &
&   Def.  & 48.6\% &   *60.0\% & .003 &    0.0\% & .003 &    0.0\% & .012 &    0.0\% & .016 &    0.0\% & .012 &    0.0\% & .011 \\
& & Rob.  & 70.6\% &    *0.0\% & .009 &    0.0\% & .010 &    0.0\% & .022 &    0.0\% & .021 &    0.0\% & .019 &    0.0\% & .019 \\
& & Sym.  & 42.2\% &   *94.7\% & .002 &   39.6\% & .002 &   85.3\% & .016 &   94.2\% & .139 &   92.8\% & .099 &   94.6\% & .102 \\
\bottomrule
\end{tabular}
\label{table:zero_attacks_test_linf}
\end{table*}

%
%
%
%
%

\textbf{Adversarial Accuracy Increases and Adversarial Robustness Decreases.} The symmetry defense increases the adversarial robustness for MILP for several datasets but decreases it for the others. The accuracy increase and robustness decrease shown in Table~\ref{table:zero_attacks_test_linf} and Table~\ref{table:zero_attacks_test_l2} in Appendix~\ref{sec:zero_attacks_test_l2} can seem counterintuitive. However, the results mean that most adversarial samples classify correctly, but those that misclassify have smaller perturbation values.

\textbf{Comparable Perturbation.} Aiming to compare accuracies for similar perturbation values, we examine whether we can adjust the adversarial attacks for the symmetry defense so that the attacks on the symmetry defense classifiers result in similar perturbation values to default and robust classifiers. However, we find that we cannot tune adversarial perturbation values.
\emph{MILP and LT-Attack} aim to find minimal perturbation, and LT-Attack only searches within a hamming distance of $1$.
\emph{Cube} also aims to find minimal perturbation by making stochastic updates near the boundary.
\emph{OPT} searches for the direction that minimizes the adversarial distortion, while \emph{SignOPT} does the same using only the gradient sign.
\emph{HSJA} determines the perturbation value with a binary search.

\subsubsection{Symmetry Defense Makes Several Attacks Unable to Find Adversarial Samples}

We find that the MILP, LT-Attack, and Cube attacks against the symmetry defense classifier do not succeed. The generated adversarial samples do not misclassify even when the adversarial sample is not inverted before classification. We show the results in Table~\ref{table:zero_attacks_no_inversion_linf} and Table~\ref{table:zero_attacks_no_inversion_l2} in Appendix~\ref{sec:non_adversarial}, and discuss them in Section~\ref{sec:discussion}.

\begin{table*}[ht!]
\caption{Accuracies of the symmetry defense classifier against zero-knowledge $L_\infty$ adversaries calculated without inverting the adversarial sample before classification. We calculate the accuracy from adversarial samples that classify correctly generated from original samples that classify correctly, out of all samples that generate correctly.}
\centering
\setlength{\tabcolsep}{0pt}
\begin{tabular}{@{\extracolsep{4pt}}lrrrrrrr@{}}  
\toprule
       &   &  \multicolumn{6}{c}{Adversarial accuracy} \\
       & Default &  \multicolumn{6}{c}{(no symmetry applied to adversarial samples)} \\
 \cmidrule{3-8}
Dataset& accuracy & MILP & LT-Attack & Cube & OPT & Sign-OPT & HSJA  \\
\midrule
breast-cancer &  87.0\% & 90.3\% & 86.0\% & 36.6\% & 0\% & 0\% & 0\% \\
diabetes      &  86.1\% & 87.3\% & 75.0\% & 56.8\% & 0\% & 0\% & 0\% \\
MNIST2-6      &  99.2\% & 78.4\% & 99.8\% & 75.8\% & 0\% & 0\% & 0\% \\
ijcnn         &  99.4\% & 91.5\% & 81.6\% & 55.7\% & 0\% & 0\% & 0\% \\
MNIST         &  98.0\% & 96.5\% & 99.6\% & 67.6\% & 0\% & 0\% & 0\% \\
F-MNIST       &  90.4\% & 96.9\% & 96.0\% & 97.8\% & 0\% & 0\% & 0\% \\
webspam       &  99.2\% & 98.4\% & 70.3\% & 93.8\% & 0\% & 0\% & 0\% \\
covtype       &  91.2\% &*93.6\% & 46.9\% & 65.4\% & 0\% & 0\% & 0\% \\
HIGGS         &  43.2\% &*63.6\% & 39.6\% & 33.8\% & 0\% & 0\% & 0\% \\
\bottomrule
\end{tabular}
\label{table:zero_attacks_no_inversion_linf}
\end{table*}

\subsection{Symmetry Defense Against Perfect-Knowledge Adversaries}

The symmetry defense against perfect-knowledge adversaries uses the invert and flip symmetries, training the XGBoost classifier for F-MNIST with original, inverted, flipped, and inverted and flipped samples.
Since perfect-knowledge adversaries are aware of the symmetry defense, the defense applies all symmetries before classification as an adversary might.
In order to classify a sample, the defense classifies the identity symmetry (original) sample, but also the inverted sample, the flipped sample, and the inverted and flipped sample. As Figure~\ref{fig:perfect} in Appendix~\ref{sec:perfect} shows, the defense decides the classification of the sample based on two classification labels of the symmetric samples that agree.
Results in Table~\ref{table:perfect_attacks_linf} for $L_\infty$ attacks and Table~\ref{table:perfect_attacks_l2} in Appendix~\ref{sec:prefect_l2} for $L_2$ attacks show that the symmetry defense exceeds default and robust classifier accuracies by up to over 95\% points.


\begin{table*}[ht!]
\caption{Here, we show accuracy and perturbation values of the XGBoost symmetry defense for F-MNIST dataset against $L_\infty$, perfect-knowledge attacks, with defense accuracies exceeding default and robust classifier accuracies by up to over 95\% points. We evaluate the symmetry defense for four cases: when perfect-knowledge adversaries generate the adversarial samples starting from original (Orig), flipped (Flip), inverted (Inv), or flipped and inverted samples (FlInv). We also compare defense accuracies with default and robust classifiers. Results against $L_2$ perfect-knowledge attacks are shown in~Table~\ref{table:perfect_attacks_linf} in Appendix~\ref{sec:prefect_l2}.}
\centering
\setlength{\tabcolsep}{0pt}
\begin{tabular}{@{\extracolsep{4pt}}lrrrrrrrrrrrrrrrrrrrr@{}}
\toprule
\multicolumn{2}{l}{Classi-} & \multicolumn{1}{c}{Default}   &  \multicolumn{12}{c}{Adversarial accuracy and perturbation} \\
\cmidrule{4-15}
\multicolumn{2}{l}{fier}    & \multicolumn{1}{c}{acc.} & \multicolumn{2}{c}{MILP} & \multicolumn{2}{c}{LT-Attack} & \multicolumn{2}{c}{Cube} & \multicolumn{2}{c}{OPT} & \multicolumn{2}{c}{SIGNOPT} & \multicolumn{2}{c}{HSJA} \\
\midrule
\multicolumn{2}{l}{Default}            & 90.4\% &    *81.0\% &*.005  &    0.0\% & .022 &    0.0\% & .036 &     0.0\% & .144 &     0.0\% & .064 &     0.0\% & .039 \\
\multicolumn{2}{l}{Robust}             & 91.2\% &    *97.7\% &*.104  &    0.0\% & .096 &    0.0\% & .094 &     0.0\% & .251 &     0.0\% & .138 &     0.0\% & .110 \\

\cmidrule{1-2} \cmidrule{3-3} \cmidrule{4-5} \cmidrule{6-7}  \cmidrule{8-9} \cmidrule{10-11} \cmidrule{12-13} \cmidrule{14-15}
\multirow{3}{*}{\rotatebox[origin=c]{90}{Defense}}
& Orig  & 91.6\% &   *100.0\% & *-    &   81.9\% & .019 &   79.7\% & .035 &    80.3\% & .150 &    84.3\% & .066 &    81.9\% & .035 \\
& Flip  & 91.6\% &   *100.0\% & *-    &   92.5\% & .005 &   96.1\% & .007 &    97.2\% & .028 &    98.3\% & .011 &    96.3\% & .019 \\
& Inv  & 91.6\% &   *100.0\% & *-    &   76.2\% & .017 &   83.2\% & .039 &    82.8\% & .178 &    86.0\% & .089 &    84.5\% & .041 \\
& FlInv & 80.0\% &    *97.4\% & *.002 &   83.5\% & .004 &   93.3\% & .005 &    96.0\% & .021 &    96.8\% & .003 &    95.5\% & .005 \\
\bottomrule
\end{tabular}
\label{table:perfect_attacks_linf}
\end{table*}

\section{Discussion of Experimental Results}\label{sec:discussion}

\textbf{Explanation for XGBoost Lack of Invariance.} Several factors contribute to the lack of invariance of XGBoost classifiers. First, the equality sign of the splitting conditions in XGBoost trees can switch from one of the branches to the other when, for example, the training dataset is inverted, as illustrated in Figure~\ref{fig:imbalance}. Even if the splitting conditions in two trees trained with original and inverted samples respectively corresponded exactly as in Figure~\ref{fig:imbalance}, samples with features equal to the splitting condition values would classify differently because the equality sign of the splitting conditions has switched branches. Feature inversion would cause branches to switch places. However, the equality sign of the condition remains on the right-hand branch for XGBoost classifiers.
Furthermore, the XGBoost split finding algorithm is greedy, could be approximate, and could also use shrinking and subsampling. Weighted quantile sketching could also result in different candidate split points. Furthermore, floating point precision could truncate feature values differently in symmetric settings. All these factors contribute to trees in symmetric settings that can classify samples differently - lack of invariance.

\textbf{Accuracy and Robustness.} The symmetry defense exceeds default and robust classifier accuracies by up to 100\% points. The defense also increases adversarial robustness values as measured by the MILP mean perturbation values for only some datasets against zero-knowledge adversaries. For $L_\infty$ attacks, adversarial robustness increases for MNIST, MNIST2-6 and covtype. For $L_2$ attacks, robustness increases for breast-cancer, diabetes, MNIST2-6 and stays the same for ijcnn and webspam. The simultaneous accuracy increase and robustness decrease for several datasets show that most adversarial samples are classified correctly, and the few that misclassify have lower perturbation.


\textbf{Training with Symmetric Samples Affects Adversarial Robustness.} Mean perturbation values generally decrease for all datasets and attacks apart from MILP. We hypothesize that training with additional symmetry samples causes the classifier to fragment the input space into smaller areas with samples with the same classification due to more splitting conditions in the classifier. More splitting conditions make the input space more fragmented and the fragments smaller, leading to smaller values of adversarial perturbation since smaller areas mean the classifier boundaries are closer to samples. To verify this, we count the number of splitting conditions in original classifiers and classifiers trained with additional symmetry samples. Table~\ref{table:fragment} in Appendix~\ref{sec:fragment} shows that for all datasets except for HIGGS, the number of splitting conditions increases from roughly 50\% to roughly 100\%, even when augmenting with only flipped images for F-MNIST classifier. We argue that fragmentation does not increase for HIGGS classifiers because of its large training dataset of $10.5M$ samples. Results in all tables support the hypothesis because the datasets are listed in increasing order of training dataset size, and corresponding perturbation values decrease. In particular, this is more pertinent when comparing corresponding perturbation values for MNIST2-6 and MNIST.

\textbf{Attacks Unable to Attack.} We make the interesting observation that the MILP, LT-Attack, and Cube attacks are unable to generate adversarial samples against the defense classifier for zero-knowledge adversaries. We hypothesize why they might not be able to.
The MILP attack is not always successful even against the default and robust classifiers for several datasets, as shown in Table~\ref{table:zero_attacks_test_linf} and Table~\ref{table:zero_attacks_test_l2} in Appendix~\ref{sec:zero_attacks_test_l2}. We think that the trees with more splitting conditions and branches make it harder for MILP to find the solution.
LT-Attack looks for adversarial samples within a hamming distance of only $1$ from the original samples. However, the increased fragmentation and branching in the classifier trees might mean that the adversarial samples are at greater hamming distances.
The Cube attack's simple algorithm that changes a subset of features might get confused by the overlapping feature values of original and inverted samples.




\textbf{Limitation.} The lack of inherent symmetries in non-image datasets impedes the application of the symmetry defense against perfect-knowledge attacks where two symmetries are needed.

\textbf{Broader impact.} The symmetry defense enables adversarial defense against several attacks without prior attack knowledge, which is important when classifiers are deployed in real-world applications without such knowledge.


\section{Conclusions}\label{sec:conclusion}


We are the first to show that XGBoost classifiers lack invariance with respect to symmetries because they can classify symmetric images differently. Without using attack knowledge, we utilize the lack of invariance of XGBoost classifiers to defend these classifiers against six attacks for adversaries that range from no knowledge to full knowledge of the defense.
Against attacks with no knowledge of the defense, we use the invert symmetry to defend even classifiers that lack inherent symmetries, exceeding default and robust classifier accuracies by up to 100\% points. Against attacks with knowledge of the defense, we use the invert and flip symmetries to exceed default and robust classifier accuracies by up to over 95\% points.
Notably, the defense makes the MILP, LT-Attack, and Cube attacks mostly unable to generate adversarial samples against classifiers trained with original and inverted samples.
Finally, we also show that legitimate symmetric samples can alter adversarial robustness.

{\small
\bibliographystyle{ieee_fullname}
\bibliography{egbib}
}

\newpage

\appendix

\section*{Appendices}

\section{Symmetry, Equivariance and Invariance}\label{sec:symmetry}

Here, we explain and define symmetry-related concepts.

\subsection{Symmetry Group Definitions} 

A group is an ordered pair $(G,*)$, where $G$ is a set and $*$ is a binary operation that acts on $G$, satisfying the following axioms~\cite{dummit2004abstract}:

\begin{itemize} [leftmargin=*]
  \setlength\itemsep{0em}
  \item \emph{Identity.} The existence of an identity element $e \in G$, such that $a*e = e*a = a$, for $\forall a \in G$.
  \item \emph{Associativity.} $*$ is associative: $\forall a,b,c \in G$, $(a*b)*c=a*(b*c)$.
  \item \emph{Inverse.} Every element of G has an inverse. This means that $\forall a \in G$, there exists $a^{-1} \in G$, such that $a*a^{-1} = a^{-1} *a = e$.
\end{itemize}

\textbf{Binary Operation.} A binary operation $*$ on a set $G$ is a function $*$: $G \times G \mapsto G$ which we can also write as $a*b$, according to~\cite{dummit2004abstract}. 

\textbf{Closure.} A subset $H$ of the $G$ set is closed under the $*$ binary operation if $\forall a,b \in H$, $a*b \in H$, according to~\cite{dummit2004abstract}.

\textbf{Group.} A group is an ordered pair $(G,*)$ of a $G$ set and a $*$ binary operation on $G$ satisfying the associativity, identity and inverse axioms, according to~\cite{dummit2004abstract}.

\textbf{Subgroup.} A subset $H$ of the $G$ set is a subgroup of $G$ if $H$ is nonempty and $H$ is closed under products and inverses (that is, $x,y \in H$ implies that $x^{-1} \in H$ and $x*y \in H$), according to~\cite{dummit2004abstract}.

\textbf{Subgroup Criterion.} A subset $H$ of a $G$ group is a subgroup if and only if $H \neq \emptyset$ and $\forall x,y \in H$, $x*y^{-1} \in H$, according to~\cite{dummit2004abstract}.

\textbf{Finite Subgroup Criterion.} An $H$ finite subset of $G$ is a subgroup if $H$ is nonempty and closed under $*$, according to~\cite{dummit2004abstract}.

\subsection{Equivariance and Invariance in CNNs.} 

A function is \textbf{equivariant} with respect to a transformation if the transformation can be equally applied to the function input or function output. Formally, we say that function $f$ is equivariant with respect to the $\mathcal{T}$ class of transformations if $\forall T \in \mathcal{T}$ of the input $x$, we can find another transformation $T ^\prime$ of the function output $f(x)$, such that $f(Tx) = T ^\prime f(x)$, based on~\cite{schmidt2012learning}. Invariance is a special case of equivariance, where the $T ^\prime$ transformation is the identity transformation. In the context of classifiers, a function that is \textbf{invariant} with respect to symmetry transformations has the same output when symmetries transformations are applied to the classifier inputs.

\section{CNN Symmetry Defense Against Perfect-Knowledge Adversaries}\label{sec:perfect}

\begin{figure}[h!]
  \centering
  \includegraphics[width=1.00\columnwidth]{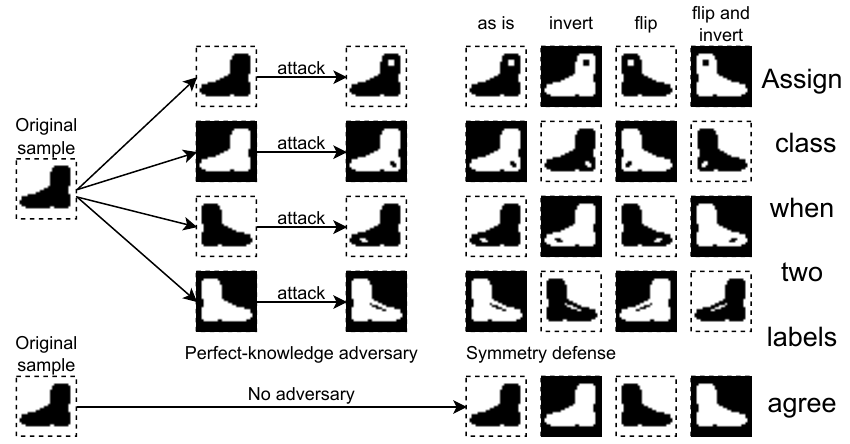}
  \caption{Here, we show that the symmetry defense against perfect-knowledge adversaries applies all four symmetry group transformations to the sample, regardless of whether the sample is adversarial. Then, the defense classifies all four symmetrically transformed samples assigning a label when there is an agreement in the classification labels of at least two symmetrically transformed samples.}
  \label{fig:perfect}
\end{figure}

Figure~\ref{fig:perfect} illustrates the symmetry defense against perfect-knowledge adversaries.

\section{The Lack of Invariance of XGBoost Classifiers}\label{sec:invariance}

Table~\ref{table:equivariance} shows that XGBoost classifiers lack invariance with respect to symmetries because the classification of original and symmetric samples can differ in a XGBoost classifier trained with both.

\begin{table}[h!]
\caption{Here, we show that GBDTs lack invariance with respect to the inversion symmetry because the classification of original and inverted testing samples can differ. For the MNIST2-6 dataset, we increase the number of testing samples to $1000$ to find a disagreement in classification.}
\label{table:equivariance}
\centering
\setlength{\tabcolsep}{4pt}
\begin{tabular}{lrrr}
\toprule
              &  & \multicolumn{2}{c}{Classification of original} \\
              & Testing & \multicolumn{2}{c}{and inverted images} \\
                        \cmidrule{3-4}
Dataset       & samples & Agree & Disagree\\
\midrule
breast-cancer & 137  & 135 & 2   \\
diabetes      & 154  & 145 & 9   \\
MNIST2-6      & 1000 & 999 & 1   \\
ijcnn         & 500  & 499 & 1   \\
MNIST         & 500  & 497 & 3   \\
F-MNIST       & 500  & 483 & 17  \\
webspam       & 500  & 494 & 6   \\
covtype       & 500  & 483 & 17  \\
HIGGS         & 500  & 150 & 350 \\
\bottomrule
\end{tabular}
\end{table}

\section{Symmetry Defense Accuracies Against Zero-Knowledge Adversaries}\label{sec:zero_attacks_test_l2}

In Table~\ref{table:zero_attacks_test_l2}, we show the defense accuracies and $L_2$ attack perturbation values for the symmetry defense against zero-knowledge adversaries.

\begin{table*}[h!]
\caption{Accuracy of the symmetry defense against $L_2$ perturbation attacks. 
Adversarial accuracies are calculated from samples that classify correctly as original samples and also as adversarial samples, out of all original samples that classify correctly. 
Adversarial perturbation mean values are calculated from samples that classify correctly as original samples and misclassify as adversarial samples out of all original samples that classify correctly.}
\label{table:zero_attacks_test_l2}
\centering

\setlength{\tabcolsep}{0pt}
\begin{tabular}{@{\extracolsep{2pt}}llllrrrrrrrrrrrrr@{}} 
\toprule
\multirow{2}{*}{\rotatebox[origin=c]{90}{Data-}} & \multirow{2}{*}{\rotatebox[origin=c]{90}{set}} & \multirow{2}{*}{\rotatebox[origin=c]{90}{Cl.}} & Def. && \multicolumn{12}{c}{Adversarial accuracy and perturbation} \\
\cmidrule{5-16}
&&& acc. & \multicolumn{2}{c}{MILP} & \multicolumn{2}{c}{LT-Attack} & \multicolumn{2}{c}{Cube} & \multicolumn{2}{c}{OPT} & \multicolumn{2}{c}{SIGNOPT} & \multicolumn{2}{c}{HSJA} \\
\midrule
\multirow{3}{*}{\rotatebox[origin=c]{90}{breast-}} & \multirow{3}{*}{\rotatebox[origin=c]{90}{cancer}}
&  Def. & 86.4\% &    0.8\% & .310 &     0.0\% & .229 &    0.0\% & .564 &    0.0\% & .373 &    0.0\% & .327 &    0.0\% & .340 \\
&& Rob. & 87.0\% &    4.5\% & .434 &     0.0\% & .434 &    0.0\% & .736 &    0.0\% & .338 &    0.0\% & .334 &    0.0\% & .336 \\
&& Sym. & 87.0\% &   93.3\% & .436 &    89.6\% & .179 &   85.1\% & .574 &   64.2\% & .165 &   91.0\% & .180 &   47.8\% & .167 \\
\cmidrule{1-3} \cmidrule{4-4} \cmidrule{5-6} \cmidrule{7-8} \cmidrule{9-10} \cmidrule{11-12} \cmidrule{13-14} \cmidrule{15-16}
\multirow{3}{*}{\rotatebox[origin=c]{90}{diabetes}} & 
&  Def. & 87.3\% &    0.0\% & .058 &    0.0\% & .311 &    0.0\% & .121 &    0.0\% & .077 &    0.0\% & .065 &    0.0\% & .069 \\
&& Rob. & 90.3\% &    0.0\% & .132 &    0.0\% & .133 &    0.0\% & .282 &    0.0\% & .110 &    0.0\% & .108 &    0.0\% & .109 \\
&& Sym. & 86.1\% &   89.3\% & .066 &   72.0\% & .057 &   76.0\% & .163 &   57.9\% & .103 &   84.3\% & .053 &   60.3\% & .081 \\
\cmidrule{1-3} \cmidrule{4-4} \cmidrule{5-6} \cmidrule{7-8} \cmidrule{9-10} \cmidrule{11-12} \cmidrule{13-14} \cmidrule{15-16}
\multirow{3}{*}{\rotatebox[origin=c]{90}{MNIST}} & \multirow{3}{*}{\rotatebox[origin=c]{90}{2-6}}
&  Def. & 99.0\% &   80.8\% & .058 &    0.0\% & .200 &    0.0\% & .610 &    0.0\% &3.169 &    0.0\% & .804 &    0.0\% & .844 \\
&& Rob. & 99.8\% &  *98.0\% &*.942 &    0.0\% & .900 &    0.0\% &1.189 &    0.0\% &4.899 &    0.0\% &1.309 &    0.0\% &1.463 \\
&& Sym. & 99.2\% &   97.4\% & .081 &   99.8\% & .002 &   96.2\% & .413 &   84.7\% & .865 &   92.7\% & .070 &   86.9\% & .050 \\
\cmidrule{1-3} \cmidrule{4-4} \cmidrule{5-6} \cmidrule{7-8} \cmidrule{9-10} \cmidrule{11-12} \cmidrule{13-14} \cmidrule{15-16}
\multirow{3}{*}{\rotatebox[origin=c]{90}{ijcnn}} &
&  Def. & 99.2\% &    0.2\% & .020 &    0.0\% & .021 &    0.0\% & .027 &    0.0\% & .026 &    0.0\% & .020 &    0.0\% & .020 \\
&& Rob. & 95.4\% &    0.0\% & .025 &    0.0\% & .026 &    0.0\% & .056 &    0.0\% & .049 &    0.0\% & .040 &    0.0\% & .040 \\
&& Sym. & 99.4\% &   82.7\% & .020 &   66.4\% & .021 &   32.3\% & .027 &   53.2\% & .028 &   68.8\% & .021 &   58.3\% & .021 & \\
\cmidrule{1-3} \cmidrule{4-4} \cmidrule{5-6} \cmidrule{7-8} \cmidrule{9-10} \cmidrule{11-12} \cmidrule{13-14} \cmidrule{15-16}
\multirow{3}{*}{\rotatebox[origin=c]{90}{MNIST}} &
&  Def. & 98.2\% &   68.6\% & .009 &    0.0\% & .048 &    0.0\% & .246 &    0.0\% &1.344 &   *0.0\% & *.251 &  *0.0\% & *.158 \\
&& Rob. & 98.6\% & *100.0\% & *-   &    0.0\% & .943 &    0.0\% &1.180 &    0.0\% &4.214 &    0.0\% & 1.329 &   0.0\% & 1.440 \\
&& Sym. & 98.0\% &   99.2\% & .005 &   99.6\% & .001 &   97.6\% & .064 &   98.0\% &*.153 &  100.0\% & *-    & 100.0\% & *-     \\
\cmidrule{1-3} \cmidrule{4-4} \cmidrule{5-6} \cmidrule{7-8} \cmidrule{9-10} \cmidrule{11-12} \cmidrule{13-14} \cmidrule{15-16}
\multirow{3}{*}{\rotatebox[origin=c]{90}{F-}} & \multirow{3}{*}{\rotatebox[origin=c]{90}{MNIST}}
&  Def. & 90.4\% &   88.5\% & .012 &    0.0\% & .054 &    0.0\% & .209 &     0.0\% &1.100 &   0.0\% & .325 &    0.0\% & .282 \\
&& Rob. & 91.2\% & *100.0\% & *-   &    0.0\% & .309 &    0.0\% & .433 &     0.0\% &1.991 &   0.0\% & .615 &    0.0\% & .631 \\
&& Sym. & 90.4\% &   98.4\% & .008 &   96.0\% & .004 &   98.2\% & .097 &    91.7\% & .919 &  97.3\% & .043 &   95.1\% & .215 \\
\cmidrule{1-3} \cmidrule{4-4} \cmidrule{5-6} \cmidrule{7-8} \cmidrule{9-10} \cmidrule{11-12} \cmidrule{13-14} \cmidrule{15-16}
\multirow{3}{*}{\rotatebox[origin=c]{90}{web-}} & \multirow{3}{*}{\rotatebox[origin=c]{90}{spam}}
&  Def. & 99.0\% &   69.9\% & .001 &   0.0\% & .003 &    0.0\% & .009 &    0.0\% & .046 &     0.0\% & .008 &   0.0\% & .018 \\
&& Rob. & 98.4\% &   *0.0\% &*.031 &   0.0\% & .041 &    0.0\% & .108 &    0.0\% & .436 &     0.0\% & .097 &   0.0\% & .132 \\
&& Sym. & 99.2\% &   99.4\% & .001 &  69.8\% & .000 &   97.1\% & .007 &   89.4\% & .062 &  *100.0\% & *-   &  96.4\% & .015 \\
\cmidrule{1-3} \cmidrule{4-4} \cmidrule{5-6} \cmidrule{7-8} \cmidrule{9-10} \cmidrule{11-12} \cmidrule{13-14} \cmidrule{15-16}
\multirow{3}{*}{\rotatebox[origin=c]{90}{covtype}} &
&  Def. & 93.0\% &  *93.6\% &*.012 &   0.0\% & .029 &    0.0\% & .060 &    0.0\% & .075 &     0.0\% & .043 &   0.0\% & .049 \\
&& Rob. & 86.0\% &  *84.4\% &*.033 &   0.0\% & .063 &    0.0\% & .164 &    0.0\% & .163 &     0.0\% & .114 &   0.0\% & .117 \\
&& Sym. & 91.2\% &  *93.6\% &*.033 &  56.3\% & .011 &   71.4\% & .064 &   70.9\% & .084 &    80.7\% & .047 &  74.4\% & .054 \\
\cmidrule{1-3} \cmidrule{4-4} \cmidrule{5-6} \cmidrule{7-8} \cmidrule{9-10} \cmidrule{11-12} \cmidrule{13-14} \cmidrule{15-16}
\multirow{3}{*}{\rotatebox[origin=c]{90}{HIGGS}} &
&  Def. & 48.6\% &    *60\% &*.006 &   0.0\% & .006 &    0.0\% & .026 &    0.0\% & .033 &     0.0\% & .017 &   0.0\% & .016 \\
&& Rob. & 70.6\% &   *0.0\% &*.014 &   0.0\% & .015 &    0.0\% & .103 &    0.0\% & .036 &     0.0\% & .023 &   0.0\% & .024 \\
&& Sym. & 42.2\% &  *97.4\% &*.003 &  39.6\% & .004 &   92.1\% & .057 &   95.0\% & .139 &   *96.2\% &*.257 &  95.3\% & .124 \\
\bottomrule
\end{tabular}
\end{table*}

\section{Some Attacks Are Not Adversarial Against The Symmetry Defense For Zero-Knowledge Adversaries}\label{sec:non_adversarial}

In Table\ref{table:zero_attacks_no_inversion_l2}, we show the accuracies of $L_2$ zero-knowledge adversaries against models that are trained with both original and inverted samples, but where we do not invert adversarial samples before classifying them. Essentially, this shows that some of the attacks cannot generate adversarial samples against classifiers with same parameters trained with original and inverted samples.

\begin{table*}[h!]
\caption{Here, we show that the MILP, LT-Attack, and Cube attacks cannot attack a classifier trained with both original and inverted samples, even when no symmetry is applied to adversarial samples after their generation. We calculate the accuracy as a percentage of generated adversarial samples that classify correctly out of all original samples that classify correctly. Here, we display $L_\infty$ attack perturbation values.}
\label{table:zero_attacks_no_inversion_l2}
\centering
\setlength{\tabcolsep}{0pt}
\begin{tabular}{@{\extracolsep{4pt}}lrrrrrrr@{}}  
\toprule
       &Default   &  \multicolumn{6}{c}{Symmetry classifier} \\
       &classifier&  \multicolumn{6}{c}{(\textbf{no symmetry applied to adversarial samples})} \\
\cmidrule{2-2} \cmidrule{3-8}
Dataset&No attack & MILP & LT-Attack & Cube & OPT & Sign-OPT & HSJA  \\
\midrule
breast-cancer &  87.0\% & 97.0\% & 89.6\% & 96.3\% & 0 & 0 & 0 \\
diabetes      &  86.1\% & 85.6\% & 72.0\% & 73.7\% & 0 & 0 & 0 \\
MNIST2-6      &  99.2\% & 97.8\% & 99.8\% & 96.2\% & 0 & 0 & 0 \\
ijcnn         &  99.4\% & 79.7\% & 66.4\% & 30.4\% & 0 & 0 & 0 \\
MNIST         &  98.0\% & 99.0\% & 99.6\% & 97.8\% & 0 & 0 & 0 \\
F-MNIST       &  90.4\% & 96.9\% & 96.0\% & 96.5\% & 0 & 0 & 0 \\
webspam       &  99.2\% & 98.0\% & 69.8\% & 95.6\% & 0 & 0 & 0 \\
covtype       &  91.2\% &*91.5\% & 56.3\% & 72.8\% & 0 & 0 & 0 \\
HIGGS         &  43.2\% &*63.6\% & 39.6\% & 33.8\% & 0 & 0 & 0 \\
\bottomrule
\end{tabular}

\end{table*}

\section{XGBoost Symmetry Defense Against $L_2$ Perfect-Knowledge Adversaries}\label{sec:prefect_l2}

In Table~\ref{table:perfect_attacks_l2}, we show the defense accuracies and $L_2$ attack perturbation values for the symmetry defense against perfect-knowledge adversaries.

\begin{table*}[h!]
\caption{Accuracy and perturbation values of the XGBoost symmetry defense for F-MNIST dataset against $L_2$, perfect-knowledge attacks. We evaluate the symmetry defense for when perfect-knowledge adversaries generate the adversarial samples starting from original, flipped, inverted, or flipped and inverted samples. We also compare with default and robust classifiers.}
\label{table:perfect_attacks_l2}
\centering
\setlength{\tabcolsep}{0pt}
\begin{tabular}{@{\extracolsep{4pt}}lrrrrrrrrrrrrrrrrrrrr@{}}
\toprule
\multicolumn{2}{l}{Classi-} & \multicolumn{1}{c}{Default}   &  \multicolumn{12}{c}{Adversarial accuracy and perturbation} \\
\cmidrule{4-15}
\multicolumn{2}{l}{fier}    & \multicolumn{1}{c}{acc.} & \multicolumn{2}{c}{MILP} & \multicolumn{2}{c}{LT-Attack} & \multicolumn{2}{c}{Cube} & \multicolumn{2}{c}{OPT} & \multicolumn{2}{c}{SIGNOPT} & \multicolumn{2}{c}{HSJA} \\
\midrule
\multicolumn{2}{l}{Default} & 90.4\% &    88.5\% & .012 &   0.0\% & .054 &    0.0\% & .037 &     0.0\% &  .132 &     0.0\% &  .088 &    *0.0\% & *.078 \\
\multicolumn{2}{l}{Robust}  & 85.8\% &  *100.0\% & *-   &   0.0\% & .309 &    0.0\% & .099 &     0.0\% &  .241 &    *0.0\% & *.176 &    *0.0\% &  .208 \\
\cmidrule{1-2} \cmidrule{3-3} \cmidrule{4-5} \cmidrule{6-7}  \cmidrule{8-9} \cmidrule{10-11} \cmidrule{12-13} \cmidrule{14-15}
\multirow{3}{*}{\rotatebox[origin=c]{90}{Defense}}
& Orig                      & 91.6\% &    99.3\% & .005 &  81.9\% & .049 &   83.0\% & .233 &    82.1\% & 1.116 &    86.9\% &  .395 &    85.2\% &  .264 \\
& Flip                      & 91.6\% &    99.6\% & .010 &  92.5\% & .011 &   97.8\% & .031 &    97.6\% &  .117 &    97.6\% &  .031 &    98.0\% &  .022 \\
& Inv                       & 91.6\% &   100.0\% & -    &  76.2\% & .047 &   85.4\% & .286 &    85.2\% & 1.407 &    87.3\% &  .481 &    86.7\% &  .418 \\
& FlInv                     & 80.0\% &    99.5\% & .003 &  83.5\% & .010 &   97.0\% & .018 &    96.3\% &  .188 &    97.3\% &  .007 &    96.5\% &  .016 \\
\bottomrule
\end{tabular}
\end{table*}

\section{Computational Resources}\label{sec:computation}

\textbf{Training.} The amount of computation for the \emph{symmetry defense against zero-knowledge adversaries} doubles because we train the dataset with a training dataset that is double the size of the original dataset due to the addition of inverted training samples.
The amount of computation for the \emph{symmetry defense against perfect-knowledge adversaries} quadruples due to the addition of inverted, flipped, and inverted and flipped training samples to the original training samples.

\textbf{Testing.} During inference for a sample, the \emph{symmetry defense against zero-knowledge adversaries} does $O(1)$ additional computation for applying the invert symmetry to the sample. The \emph{symmetry defense against perfect-knowledge adversaries} does $O(1)$ additional computation for applying the four symmetries to the sample and for choosing the label from two classification labels that agree, as well as quadruples the amount of computation for the inference of a sample.

\section{More Fragmentation of The Input Space When Training With Both Original and Symmetric Samples}\label{sec:fragment}

In Table~\ref{table:fragment}, we show that training with both original and inverted samples causes the input space to fragment into smaller chunks. These smaller chunks cause smaller perturbation values because it means that chunks of the input space with adversarial samples are closer. Only the HIGGS dataset does not result in more fragmentation, which can be explained by the big size of $10.5M$ of the HIGGS training dataset. Due to the big HIGGS dataset training dataset size compared to other datasets (shown in Table~\ref{table:params}), the inclusion of inverted samples in the training set does not introduce more fragmentation of the input space. Instead, for HIGGS dataset, the extra inverted samples decrease the fragmentation, possibly because the feature values in inverted samples fill out gaps in the distribution of feature values.

\begin{table}[ht!]
\caption{Here, we show that by training with additional symmetric samples, the input space is much more fragmented than in default classifiers, leading to smaller adversarial perturbation values. To show the bigger fragmentation of the input space into smaller chunks, we compare the number of splitting conditions in default classifier models and in classifier models trained with the same parameters but with additional symmetric samples. The only exception is the HIGGS dataset, which we explain with the big size of $10.5M$ of the HIGGS training dataset.}
\centering
\setlength{\tabcolsep}{0pt}
\begin{tabular}{@{\extracolsep{3pt}}lrrrr@{}}
\toprule
       & \multicolumn{4}{c}{Number of tree splitting conditions} \\
              \cmidrule{2-5}
              &              & \multicolumn{3}{c}{Classifer trained with}    \\
              \cmidrule{3-5}
              &              & orig. & orig.   & orig.    \\
              &              & inv.  & flipped & flipped  \\
              &   Default    &       &         & inverted \\
 Dataset      &   classifier &       &         & fl. and inv. \\
              \midrule
breast-cancer &     72 &    \textbf{152} \\
diabetes      &    282 &    \textbf{394} \\
MNIST2-6 &  10666 &  \textbf{14503} \\
ijcnn         &   4270 &   \textbf{7367} \\
MNIST         &  60286 & \textbf{117947} \\
Fashion-MNIST &  93095 &         162830  & 144634 & \textbf{163775} \\
webspam       &  11875 &  \textbf{15119} \\
covtype       & 127193 & \textbf{174825} \\
HIGGS         & \textbf{73613} &  70796 \\
\bottomrule
\end{tabular}
\label{table:fragment}
\end{table}

\section{Training Parameters}\label{app:training_params}

Here, we reference the default training parameters we used for the defense and the parameters for the robust classifiers we used for comparison. The parameter values and the trained robust classifiers were obtained from~\cite{chen2019robustness}. 


\begin{table*}[ht!]
\caption{Training parameters for XGBoost models, obtained from~\cite{chen2019robustness}.}
\label{table:params}
\centering
\setlength{\tabcolsep}{3pt}
\begin{tabular}{lrrrrrrrrrrr}
\toprule

Dataset       &   Training & Testing  & No. of & No. of & No.   & Robust & \multicolumn{2}{c}{Depth} & \multicolumn{2}{c}{Stand. test acc.} \\
              &   set size & set size & features   & classes   & of trees &   eps  &Rob. & Nat. &  robu &natu \\
\midrule
breast-cancer &        546 &      137 &   10 &    2   &    4 &   0.3  & 8 & 6 &  97.8\% & 96.4\% \\
diabetes      &        614 &      154 &    8 &    2   &   20 &   0.2  & 5 & 5 &  78.6\% & 77.3\% \\
MNIST2-6      &     11,876 &    1,990 &  784 &    2   & 1000 &   0.3  & 6 & 4 &  99.7\% & 99.8\% \\
ijcnn         &     49,990 &   91,701 &   22 &    2   &   60 &   0.1  & 8 & 8 &  95.9\% & 98.0\% \\
MNIST         &     60,000 &   10,000 &  784 &   10   &  200 &   0.3  & 8 & 8 &  98.0\% & 98.0\% \\
F-MNIST       &     60,000 &   10,000 &  784 &   10   &  200 &   0.1  & 8 & 8 &  90.3\% & 90.3\% \\
webspam       &    300,000 &   50,000 &  254 &    2   &  100 &   0.05 & 8 & 8 &  98.3\% & 99.2\% \\
covtype       &    400,000 &  181,000 &   54 &    7   &   80 &   0.2  & 8 & 8 &  84.7\% & 87.7\% \\
HIGGS         & 10,500,000 &  500,000 &   28 &    2   &  300 &   0.05 & 8 & 8 &  70.9\% & 76.0\% \\
\bottomrule
\end{tabular}
\end{table*}

\end{document}